\documentclass[letterpaper, 10 pt, conference]{ieeeconf}
\IEEEoverridecommandlockouts
\usepackage{graphics} 
\usepackage{epsfig} 
\usepackage{amsmath} 
\usepackage{amssymb}  
\usepackage{graphicx}
\usepackage{booktabs} 
\usepackage{cite}
\usepackage{algorithm}
\usepackage{algorithmic}
\usepackage{xcolor}
\usepackage{float}
\usepackage{placeins}    
\usepackage{balance}    
\usepackage{emptypage} 
\usepackage{etoolbox}
\AtBeginEnvironment{thebibliography}{\footnotesize}  
\makeatletter
\@ifundefined{bibitemsep}{\newlength{\bibitemsep}}{}
\makeatother

\title{\LARGE \bf
The Moving Eye: Enhancing VLA Spatial Generalization via \\ Hybrid Dynamic Data Collection

}

\author{%
  Jincheng Tang$^{1\dagger}$, Yilong Zhu$^{1,2\dagger}$, Zhengyuan Xie$^{1,3\dagger}$, Jiang-Jiang Liu$^{1*}$, Jiaxing Zhang$^{1}$%
  \thanks{$^{\dagger}$ Equal contribution.}%
  \thanks{$^{*}$ Corresponding author and project leader: Jiang-Jiang Liu (\texttt{j04.liu@gmail.com}).}%
  \thanks{$^{1}$ Lion Rock AI Lab, China Merchants Research Institute of Advanced Technology.}%
  \thanks{$^{2}$ The Hong Kong University of Science and Technology.}%
  \thanks{$^{3}$ Nankai University.}%
}

\begin{document}

\maketitle
\thispagestyle{empty}
\pagestyle{empty}

\begin{abstract}

Vision-Language-Action (VLA) models have shown remarkable promise in generalized robotic manipulation. However, their spatial generalization remains fragile. We argue that simply increasing the number of viewpoints is insufficient. Models often fall into the trap of Shortcut Learning, latching onto spurious correlations (e.g., fixed relative poses between objects or between the camera and robot base) rather than learning true spatial relationships. In this work, we propose a data-centric solution to enhance VLA spatial generalization. We utilize a dual-arm setup where one arm performs manipulation while the other serves as a mobile environmental camera. We systematically evaluate three data distribution patterns: Fixed, Multi-Fixed, and Moving Views. Our findings reveal that a hybrid strategy—combining continuous camera motion with diverse static viewpoints—yields the best performance by substantially reducing spurious correlations while maintaining training stability. Our experiments demonstrate that this strategy mitigates spurious correlations, enabling VLAs to generalize to unseen camera poses and object configurations where simply adding more static viewpoints fails. Crucially, we reveal that the susceptibility to shortcut learning and the struggle with spatial generalization are universal characteristics shared across diverse architectures. Consequently, all evaluated models (ACT, Diffusion, and VLA models including Pi0 and Gr00t) benefit significantly from our mixed data strategy.

\end{abstract}

\section{INTRODUCTION}

Vision-Language-Action (VLA) models have demonstrated impressive semantic understanding and manipulation capabilities. However, their spatial generalization remains fragile: slight perturbations in camera pose or object configuration can expose substantial robustness and generalization gaps \cite{fei2025liberoplus,zhou2025liberopro,xie2023decomposing,chen2026radar}. Current methods struggle to maintain robustness when the visual input deviates even marginally from the training distribution.

We identify \textbf{shortcut learning} as a key contributing factor: models often exploit spurious camera--robot--object regularities instead of learning task-relevant spatial relations. We categorize these shortcuts into three types of implicit couplings:
\begin{itemize}
    \item \textbf{Camera-Base Coupling:} The model memorizes the robot's static appearance relative to the background (cf. Exp.~1).
    \item \textbf{Camera-Object Coupling:} The model relies on fixed camera angles to recognize objects, failing when viewed from new perspectives.
    \item \textbf{Object-Position Coupling:} The model overfits to the fixed relative positions between objects (e.g., a pen and a holder). As we demonstrate in Exp.~2, even with diverse camera views, if the relative position of the target and receptacle remains static, the model exhibits a clear drop in generalization to new configurations.
\end{itemize}

\begin{figure}[t]
  \centering
  \includegraphics[width=0.95\columnwidth]{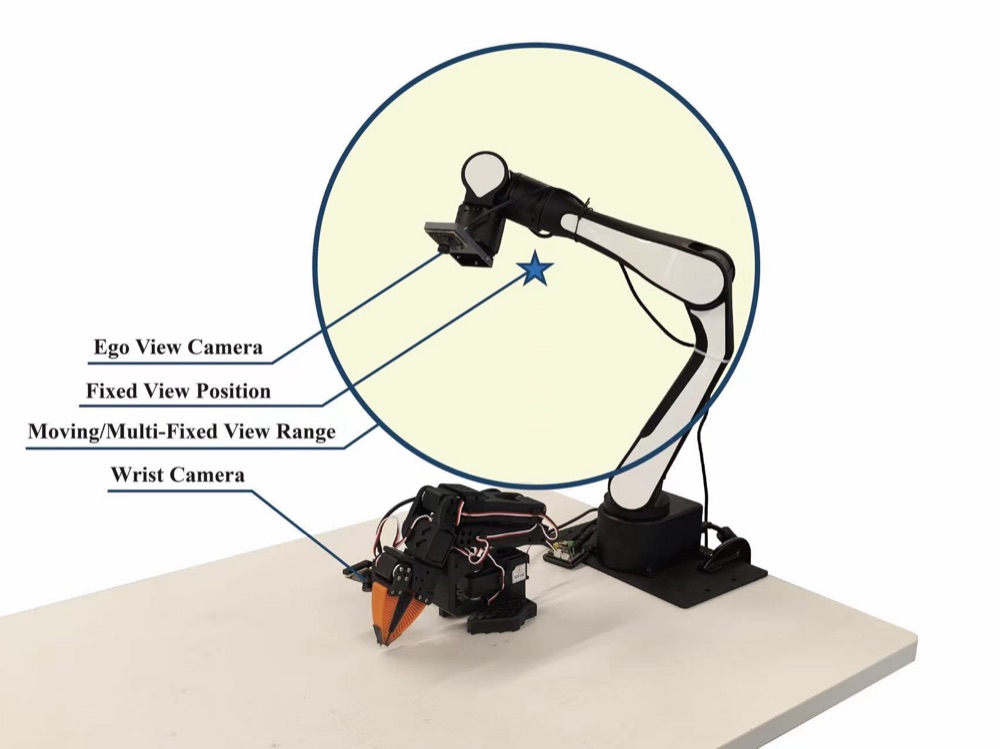}
  \caption{\textbf{Hierarchical Spatial Zoning for Data Collection.} We conceptualize the robot's camera workspace into three configurations: \textbf{Fixed View}, where the camera is static; \textbf{Multi-Fixed View}, where the camera is static within an episode but varies across episodes; and \textbf{Moving View}, the decoupling zone where the camera moves continuously to break spurious correlations. Our key insight is that low-cost decoupled data from Moving Views can robustify the policy trained on Multi-Fixed Views.}
  \label{fig:teaser}
\end{figure}
Addressing these couplings is critical for real-world deployment, where visual stability is rarely guaranteed. As summarized in Table \ref{tab:view_definitions}, we categorize camera configurations into three types based on their motion characteristics. While Fixed and Multi-Fixed views are common, they are susceptible to the aforementioned couplings. Real-world scenarios often demand robustness across these configurations:
\begin{itemize}
    \item \textbf{Static Cameras (Fixed View):} Even fixed cameras suffer from mechanical vibrations or accidental nudges.
    \item \textbf{Fixed Multi-View Scenarios (Multi-Fixed View):} Multi-Fixed View typically arises when data are collected from different camera rig positions, when multiple datasets with different ego-camera placements are mixed for training, or when using surveillance requiring robustness across discrete viewpoints.
    \item \textbf{Moving View Scenarios (Moving View):} Dynamic scenarios such as VR/AR or ego-centric data \cite{engel2023project}, handheld operation, and mobile manipulation demand continuous spatial understanding.
\end{itemize}

\begin{table}[h]
\centering
\caption{\textbf{Experimental Conditions: View Definitions.} Summary of the three camera configurations used in our experiments.}
\label{tab:view_definitions}
\resizebox{\columnwidth}{!}{%
\begin{tabular}{lccc}
\toprule
\textbf{View Type} & \textbf{Intra-Episode Motion} & \textbf{Dataset Distribution} & \textbf{Spatial Range} \\
\midrule
Fixed View & Static & Single Fixed Pose & Point \\
Multi-Fixed View & Static & Discrete Poses & Bounded Region \\
Moving View & Dynamic & Continuous Trajectories & Bounded Region \\
\bottomrule
\end{tabular}%
}
\end{table}

To tackle this, we propose a \textbf{Hierarchical Data Decoupling} strategy. We implement a real-world dynamic camera framework that not only employs a hybrid strategy of \textbf{Multi-Fixed} and \textbf{Moving Views} to break Camera-Base and Camera-Object couplings but also systematically injects \textbf{Multi-dimensional Diversity} into the data collection process. Specifically, we explicitly vary the relative positions of objects (e.g., the pen holder) during data collection to break Object-Position coupling, ensuring the model learns robust spatial representations.

Our contributions are:
\begin{itemize}
    \item \textbf{Real-World Realization:} We present a systematic real-world validation of a dynamic viewpoint data collection system, bridging the gap between simulation-based paradigms (e.g., MOVE \cite{move2024}) and physical deployment.
    \item \textbf{Shortcut Breaking:} We demonstrate that our Hybrid Dynamic Data Collection strategy effectively mitigates spurious correlations (e.g., Object-Position Coupling) that naive augmentation schemes cannot resolve.
    \item \textbf{Transfer \& Efficiency:} Moreover, we show that the decoupled dynamic data learned from an auxiliary task can be efficiently transferred to enhance the robustness of static policies on unseen tasks, enabling high-performance generalization with only a small amount of data.
\end{itemize}

\section{RELATED WORK}

\subsection{Manipulation Viewpoint Robustness}

Recent Vision--Language--Action (VLA) models~\cite{bjorck2025gr00t,intelligence2025pi05visionlanguageactionmodelopenworld} achieve strong performance on manipulation benchmarks but remain fragile under camera viewpoint shifts. Improving robustness is therefore closely tied to learning view-invariant, geometry-aware, and action-relevant representations.

One line of work incorporates explicit 3D or depth cues. SPA~\cite{zhu2024spa} pretrains a ViT with 3D spatial awareness via differentiable neural rendering on multi-view images. GeoAware-VLA~\cite{abouzeid2025geoaware} improves viewpoint invariance by combining a frozen geometry-aware encoder with a lightweight projection to the policy decoder. OG-VLA~\cite{singh2025og} renders multi-view RGB-D into canonical orthographic views and predicts 6-DoF keyframes via LLM-conditioned diffusion, enabling few-shot 3D generalization. RVT~\cite{goyal2023rvt} maps RGB-D to multi-view virtual images and uses a Transformer to predict 3D keyframe actions efficiently. ManiVID-3D~\cite{li2026manivid} further learns view-invariant RL policies via cross-view point cloud alignment and contrastive learning.

Another line explicitly injects camera information into the perception--policy pipeline to disentangle viewpoint from geometry. FTM~\cite{li2025vla} adapts policies via one-shot visual feature modulation. Acar et al.~\cite{acar2023visual} distill multi-view teachers trained with camera randomization into single-view students for better robustness. Jiang et al.~\cite{jiang2025you} encode camera extrinsics using per-pixel Plücker embeddings, reducing viewpoint-induced failures. Vantage~\cite{vasudevan2025agnostic} uses Bayesian optimization to select informative camera poses for efficient viewpoint-agnostic fine-tuning.

A third line leverages self-supervised learning to enforce cross-view consistency. MV-MWM~\cite{seo2023multi} learns view-masked representations and trains a world model for robust transfer. CL3R~\cite{cui2025cl3r} combines point-cloud MAE with contrastive alignment to 2D models while unifying multi-view geometry via extrinsics.

Despite these advances, simply increasing viewpoint diversity can still induce shortcut learning. We therefore move toward a data-centric paradigm that explicitly reshapes view distributions to mitigate spurious correlations and improve spatial generalization.
\subsection{Data Paradigm for Manipulation Viewpoint Robustness} 
Beyond architecture-level interventions, an emerging view is that viewpoint robustness can be substantially improved by data-centric design.
Some prior works use data augmentation to expand the effective support of camera viewpoints, synthesizing counterfactual observations under novel poses so that the policy learns to focus on task-relevant geometry rather than viewpoint-specific appearance cues.
VISTA~\cite{tian2024vista} improves cross-view robustness by augmenting single-view demonstrations with zero-shot diffusion-based novel-view synthesis during policy training.
RoboSplat~\cite{yang2025novel} generates diverse, spatially consistent manipulation demonstrations by editing a 3D Gaussian-splatting reconstruction (including novel-view synthesis), substantially improving real-world generalization from a single demonstration.
Invariance Co-training~\cite{yang2025invariance} boosts viewpoint robustness by co-training on real demos and cheap non-physics simulated images with contrastive invariance and camera-extrinsics regression. 
Another emerging direction is to compose training data by strategically selecting, mixing, and reweighting viewpoints/episodes to form a more informative and balanced multi-view curriculum.
ADC~\cite{huang2025adversarial} collects denser demonstrations by injecting real-time visual and language perturbations during teleoperation, improving compositional generalization and robustness with fewer trajectories.
MOVE~\cite{move2024} suggests motion-based collection but leaves real-world deployment as future work. 
We extend this by implementing it on a real robot and adding the critical dimension of Object-Position decoupling to prevent overfitting.

\subsection{Active Perception vs.\ Data-Centric Decoupling}
A complementary line learns to \emph{control} the viewpoint online: Vision-in-Action~\cite{xiong2025vision} and ActiveUMI~\cite{zeng2025activeumi} learn task-oriented head or gaze motions from human demonstrations to resolve occlusion and improve observability, echoing classical next-best-view planning. Our objective is orthogonal and complementary: rather than optimizing \emph{where} to look at test time, we shape the \emph{training-data distribution} so that the policy becomes viewpoint-invariant, removing the camera--base and object--position shortcuts that persist under any viewing policy. The paradigms compose---an active-perception controller can sit atop a policy that our task-agnostic, broadly sampled moving-view data has already made viewpoint-robust, and we expect the same data to improve observability under occlusion as a side effect, though we do not evaluate this separately.

\section{METHODOLOGY}

\begin{figure*}[t]
    \centering
    \includegraphics[width=\textwidth]{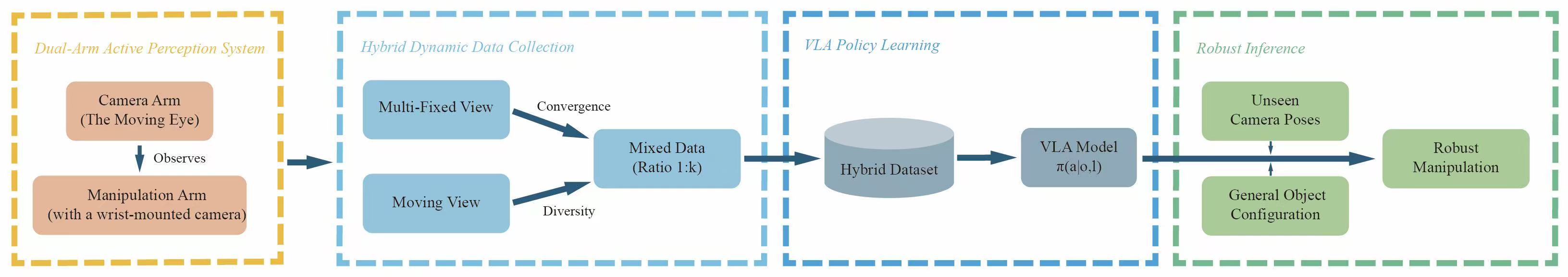}
    \caption{\textbf{System pipeline.} (1) \textit{Dual-arm:} The manipulation arm (with wrist camera) executes actions; the environmental camera arm observes the scene from varying viewpoints. (2) \textit{Hybrid Dynamic Data Collection:} Multi-Fixed View data provide stability for convergence, Moving View data provide decoupling to break spurious correlations; they are mixed at ratio Moving:Multi-Fixed $= 1:k$ (e.g., $k=3$ for the Golden Ratio) to form $\mathcal{D}_{train}$. (3) \textit{VLA policy learning:} The hybrid dataset trains a VLA policy $\pi(a|o,l)$ mapping observations and language to actions. (4) \textit{Deployment:} The policy is evaluated on unseen camera poses and object configurations for robust spatial generalization.}
    \label{fig:pipeline}
\end{figure*}

We aim to break shortcut learning via a data collection strategy that is realizable (Fixed, Multi-Fixed, Moving) and maximizes diversity within that framework. Fig.~\ref{fig:pipeline} summarizes the overall system pipeline.

\subsection{Problem Setup}
We consider a VLA policy $\pi(a|o,l)$ that predicts action $a$ given observation $o$ (visual input from the wrist and environmental cameras) and language instruction $l$. The policy takes input from a wrist-mounted camera on the manipulation arm and a mobile environmental camera controlled by another robot arm. The pose $P_c$ of the environmental camera lies within a workspace $\mathcal{W}$.

\subsection{Camera Viewpoint Configurations}
We define three distinct camera viewpoint configurations to systematically vary the data distribution, as illustrated in Fig. \ref{fig:motion_range}:

\begin{figure}[h]
    \centering
    \includegraphics[width=0.95\columnwidth]{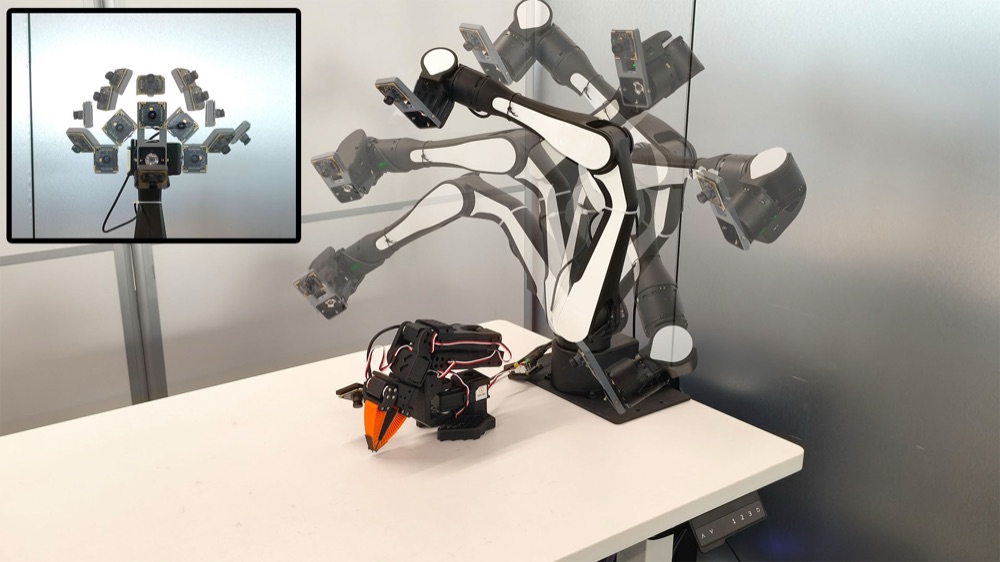}
    \caption{\textbf{Real-world Realization of Camera Viewpoint Configurations.} We implement Fixed, Multi-Fixed, and Moving Views using an environmental camera controlled by a separate robot arm. The Moving View configuration (Decoupling Zone) allows for continuous trajectory sampling to break spurious correlations.}
    \label{fig:motion_range}
\end{figure}

\begin{itemize}
    \item \textbf{Fixed View:} The camera remains at a single static pose throughout all episodes. This represents the standard, constrained setting.
    \item \textbf{Multi-Fixed View:} The camera pose is static within each episode but varies discretely across episodes within a bounded region. This introduces viewpoint diversity without intra-episode motion.
    \item \textbf{Moving View:} The camera moves continuously along trajectories within a bounded region (same as the region of Multi-Fixed View) during data collection. This decoupling zone maximizes the variation in relative poses between the camera, robot, and objects.
\end{itemize}
We hypothesize that the Moving View configuration, by providing continuous viewpoint changes, is the most effective at breaking spurious correlations.

\subsection{Multi-dimensional Diversity Injection}
To prevent shortcut learning, we enforce diversity along three axes during collection (consistent with the data collection settings in the experiment summary):
\begin{enumerate}
    \item \textbf{Viewpoint Diversity:} Sampling $P_c$ across Multi-Fixed and Moving configurations (random per-group camera pose and per-episode moving trajectories in our setup).
    \item \textbf{Object Configuration Diversity:} Randomizing relative poses between target and receptacle (e.g., pen and holder at diverse positions in general collection).
    \item \textbf{Base-Camera Decoupling:} Ensuring no fixed relationship between the camera frame and the robot base frame via viewpoint variation (environmental camera motion in our experiments, not moving the robot base).
\end{enumerate}

\subsection{Hybrid Dynamic Data Collection Strategy}
To effectively break the aforementioned couplings, we propose a \textbf{Hybrid Dynamic Data Collection} strategy. As detailed in the experiment summary, this strategy integrates three key components:
\begin{enumerate}
    \item \textbf{Hierarchical Viewpoint Sampling:} We collect data across Fixed, Multi-Fixed, and Moving View configurations. While Multi-Fixed data provides stability for convergence, Moving View data acts as a regularizer to enforce spatial invariance.
    \item \textbf{Multi-dimensional Diversity Injection:} Beyond camera motion (provided by hierarchical viewpoint sampling), we explicitly vary the relative positions of objects (e.g., target and receptacle) to break Object-Position coupling. Camera-Base decoupling is achieved by viewpoint variation (Moving and Multi-Fixed); we do not explicitly vary the robot base in our experiments.
    \item \textbf{Optimal Composition:} We investigate the mixing ratio between Multi-Fixed and Moving data. Let $k$ denote the Multi-Fixed part in the ratio Moving:Multi-Fixed $= 1:k$. The mixture is
    \begin{equation}
        \mathcal{D}_{train} = \tfrac{k}{k+1} \mathcal{D}_{MultiFixed} + \tfrac{1}{k+1} \mathcal{D}_{Moving}
    \end{equation}
    Our experiments reveal an empirically optimal mixing ratio for our primary model (Gr00t~n1, hereafter Gr00t): \textbf{Moving:Multi-Fixed = 1:3} (we refer to it as the Golden Ratio for brevity). Exp.~4 shows the best $k$ can differ per architecture.
\end{enumerate}

\begin{algorithm}[h]
\caption{Hybrid Dynamic Data Collection Strategy}
\label{alg:hybrid_collection}
\begin{algorithmic}[1]
\REQUIRE Task $\mathcal{T}$, Workspace $\mathcal{W}$, ratio Moving:Multi-Fixed $= 1:k$ ($k$ = Multi-Fixed part, e.g.\ $k=3$ for Golden Ratio)
\ENSURE Hybrid Dataset $\mathcal{D}_{train}$
\STATE Initialize $\mathcal{D}_{train} \leftarrow \emptyset$  
\FOR{episode $e = 1$ to $N$}  %
    \STATE Sample random number $r \sim U(0, 1)$ 
    \IF{$r < 1/(k+1)$}  
        \STATE \textbf{Mode: Moving View} (Decoupling)  
        \STATE Sample trajectory $\tau_c(t)$ within $\mathcal{W}_{bounded}$  
        \STATE Move Camera Arm along $\tau_c(t)$ continuously  
    \ELSE  
        \STATE \textbf{Mode: Multi-Fixed View} (Stability)  
        \STATE Sample static camera pose $P_c \sim \text{Uniform}(\mathcal{W}_{bounded})$  
        \STATE Move Camera Arm to $P_c$ and hold 
    \ENDIF
    \STATE Randomize object configuration $P_{obj}$ (Object Diversity)  
    \STATE Collect trajectory data $\tau_{data} = \{(o_t, a_t, P_c^t)\}_{t=0}^T$  
    \STATE $\mathcal{D}_{train} \leftarrow \mathcal{D}_{train} \cup \tau_{data}$  
\ENDFOR
\RETURN $\mathcal{D}_{train}$  
\end{algorithmic}
\end{algorithm}

\section{EXPERIMENTS}

\subsection{Experimental Setup}
\begin{figure}[h]
    \centering
    \includegraphics[width=0.95\columnwidth]{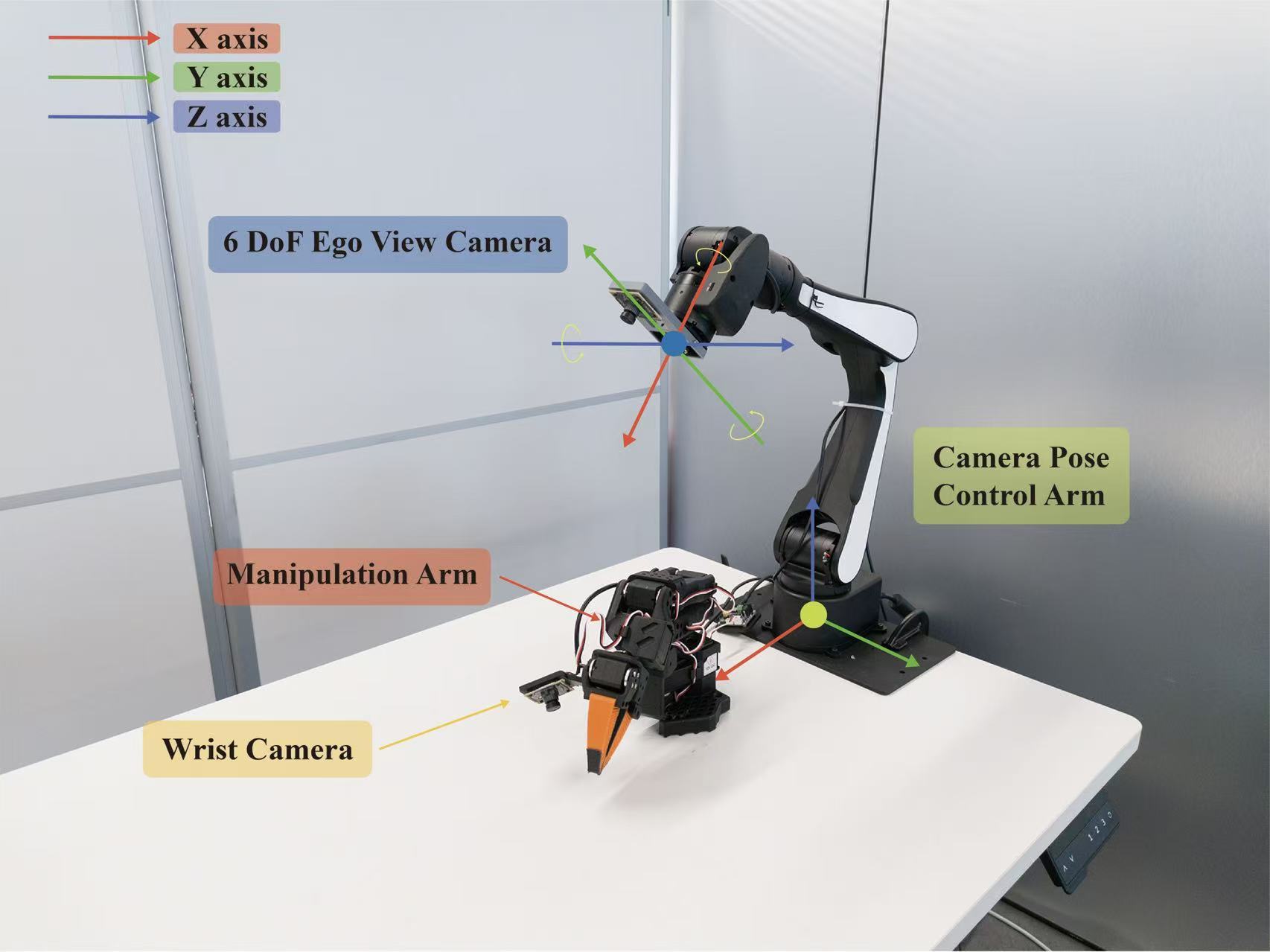}
    \caption{\textbf{Experimental Setup.} A robot arm controls a mobile environmental eye capturing data from various viewpoints, while another one with a wrist-mounted camera performs manipulation tasks.}
    \label{fig:setup}
\end{figure}

The experimental setup is shown in Fig. \ref{fig:setup}.

\begin{itemize}
    \item \textbf{Platform:} So-101 robot arm with a wrist-mounted camera for manipulation and an Airbot arm controlling the environmental eye. Images are captured with WowRobo 2\,MP USB camera modules at $640\times480$ resolution and 30\,FPS.
    \item \textbf{Task:} Pick-and-place spanning a pen task (grasping a pen and \emph{inserting} it into a holder, a contact-rich alignment beyond free-space placement) and a five-category multi-object task (watch, lollipop, nail clipper, tape, cube) covering varied geometry and affordances.
    \item \textbf{Sensing \& Camera Motion:} For Moving View collection, the environmental camera is commanded in end-effector space along continuous trajectories at a constant linear speed of $0.05$\,m/s, with a mean viewpoint angular speed of $0.198$\,rad/s (max $0.419$\,rad/s). All observations are recorded as continuous MP4 video at 30\,FPS through the open-source LeRobot SO-101 data-collection pipeline, rather than as discrete JPEG/PNG snapshots.
    \item \textbf{Baselines:}
    \begin{enumerate}
        \item \textbf{Baseline (Fixed View):} Standard practice with a single static camera pose.
        \item \textbf{Baseline (Multi-Fixed View):} Discrete static viewpoints, prone to shortcut learning.
        \item \textbf{Ours (Mixed Data):} Hybrid strategy combining Multi-Fixed and Moving Views with diversity injection.
    \end{enumerate}
\end{itemize}

\textbf{Data Collection Settings:}
\begin{itemize}
    \item \textbf{Pen Task (Multi-Fixed for Coupling Test):} Collected in groups of 5 episodes. In each episode, 1 to 5 pens are grasped sequentially. Within a group, the camera pose is randomly generated but remains fixed across all 5 episodes, and the pen holder is kept at a fixed position to simulate coupling.
    \item \textbf{Pen Task (General):} Similar to above, but for general Multi-Fixed data and Moving data, the pen and the pen holder are placed at diverse positions. For Moving data, the camera moves continuously along random trajectories in every episode.
    \item \textbf{Multi-Task:} Collected in groups of 5 episodes, each grasping a different object (watch, lollipop, nail clipper, tape, cube block), with the target object and receptacle placed at diverse positions. All data use a single Fixed View (not Multi-Fixed).
    \item \textbf{Note:} Pen Task episodes involve multiple grasps (1--5) so the model captures viewpoint changes during slow camera motion. Except for the coupling-test data, we maximized diversity (object poses, instances, orientations) to break potential spurious correlations.
\end{itemize}

\textbf{Evaluation Settings:}
We generated a fixed set of 40 camera moving trajectories (for Moving Test) and 40 camera poses (for Multi-Fixed Test). We also defined 5 fixed target-container relative positions and 8 pen orientations ($0^\circ, 45^\circ, ..., 315^\circ$).
\begin{itemize}
    \item \textbf{Moving Test:} In the $i$-th try, the camera follows the $i$-th trajectory from the list.
    \item \textbf{Multi-Fixed Test:} In the $i$-th try, the camera is set to the $i$-th pose from the list.
    \item \textbf{Fixed Test:} The camera is set to the predefined fixed pose.
\end{itemize}
For Pen tasks, each group consists of 40 pick-and-place tries (1 pen per try). For Multi-Task, each group consists of 25 tries (5 tries per object type). \textbf{Success rate computation:} Reported success rates are computed over \textbf{400 evaluation episodes} for Pen tasks and over \textbf{100 episodes} for Multi-Task (20 tries per object type $\times$ 5 object types). \textbf{ID/OOD convention:} We use in-distribution (ID) and out-of-distribution (OOD) consistently per experiment: in Exp.1, ID = fixed camera and holder as in training, OOD = moving camera; in Exp.2, ID = pen and pen holder at training position, OOD = holder shifted by one diameter (pen remains at the same position as in ID).

\textbf{Training Details:} All Gr00t \cite{bjorck2025gr00t} models were trained on 8 NVIDIA H800 GPUs. We used a batch size of 4 and an initial learning rate of 1e-4 with a cosine decay schedule. Training for the Pen Pick-and-Place task (2400 episodes) took approximately 34--37 hours, while the Multi-Task experiment (2400 episodes) took approximately 14 hours---the Pen episodes' multiple grasps (1--5) yield $\sim$2.5$\times$ more frames.

\textbf{Why does Moving-View work?} In our opinion, Moving-View robustness stems from \emph{dense, near-uniform multi-view sampling}, which we expect to hold across moderate speeds; since frame rate mainly sets the effective data density per trajectory, lowering it behaves like reducing data---success should degrade gracefully and the gap may vanish given enough episodes. What breaks this is a large inter-frame viewpoint jump (excessive speed $v$ or speed-to-frame-rate ratio): as in Exp.~1, a Fixed-View policy under moving viewpoints misjudges its egocentric pose and mis-positions at pick/place ($85\%\!\to\!43\%$), so we consider stable manipulation under arbitrarily fast or abrupt viewpoint motion to be out of scope.

\subsection{Exp. 1: Shortcut Learning (Camera-Base Coupling)}
We first evaluate whether models exhibit shortcut learning. We train on a \textbf{Fixed View} dataset (camera static across all episodes) and test in two settings:
\begin{itemize}
    \item \textbf{ID-Test (Fixed):} Evaluation on the same static viewpoint as training. 
    \item \textbf{OOD-Test (Moving):} Evaluation on dynamic, continuous viewpoint trajectories (Moving View).  
\end{itemize}

\begin{figure}[ht]
    \centering
    \includegraphics[width=0.95\columnwidth]{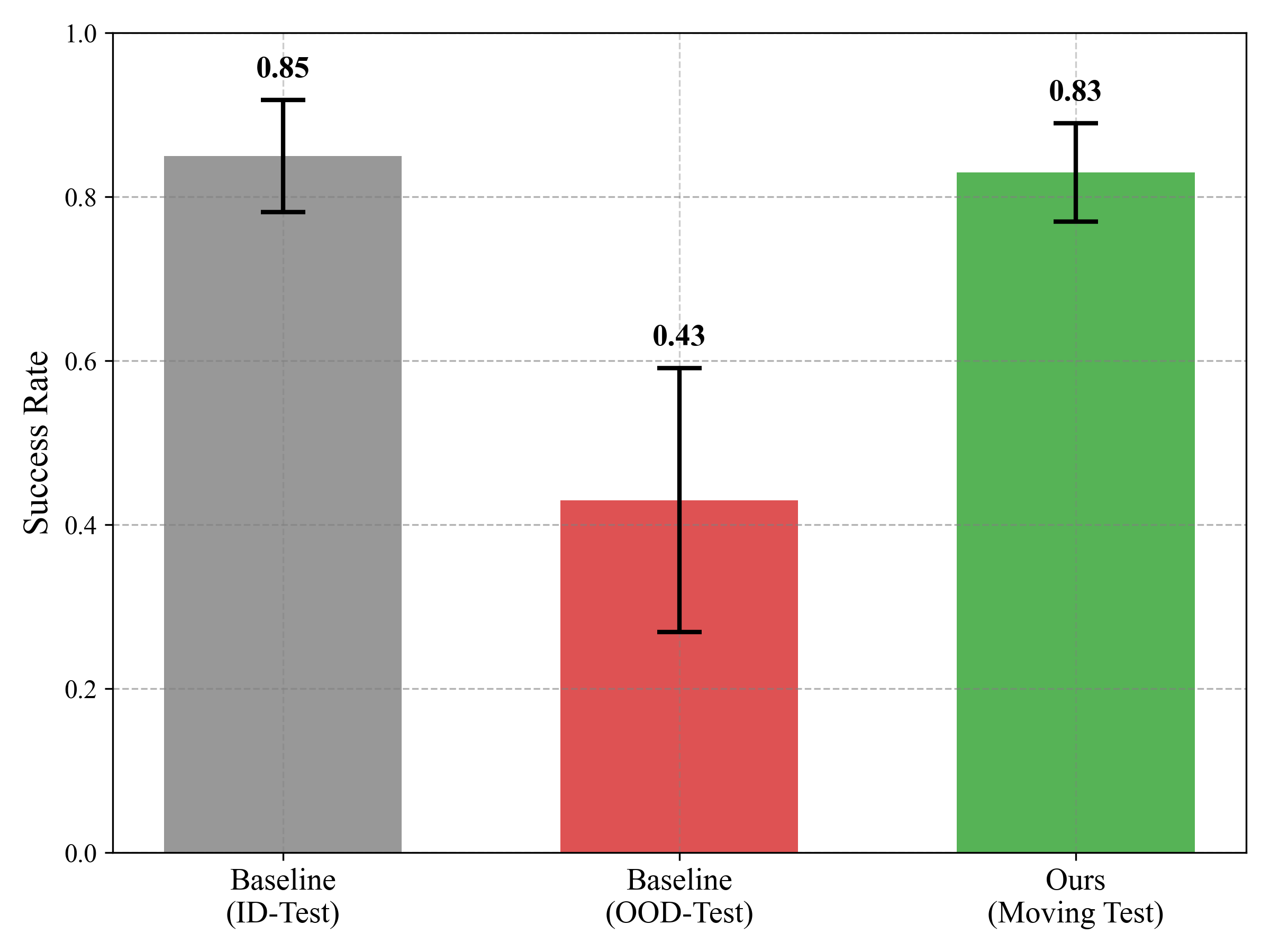}
    \caption{\textbf{Shortcut learning (Camera-Base).} While the Fixed View Baseline collapses on Moving Tests (OOD), our method maintains high performance. Error bars denote standard deviation. Here, ID = same fixed camera and holder as training; OOD = moving camera (Exp.1 convention).}
    \label{fig:shortcut}
\end{figure}

As shown in Table \ref{tab:shortcut} and Fig. \ref{fig:shortcut}, the Baseline (Fixed View Data) achieves high success (85\%) on ID-Test but collapses to 43\% on OOD-Test (Moving), revealing severe overfitting. Our method (Mixed Data) maintains high robustness (83\%) even under dynamic viewpoints (see Exp.~1 ID-Test for fixed-view retention).

\begin{table}[h]
\centering
\caption{\textbf{Breaking the Shortcut.} Success rates (\%) at 2400 samples. The gap between ID and OOD reveals shortcut learning.}
\label{tab:shortcut}
\footnotesize
\begin{tabular}{lcc}
\toprule
\textbf{Method} & \textbf{ID-Test (Fixed)} & \textbf{OOD-Test (Moving)} \\
\midrule
Baseline (Fixed) & 85.0 & 43.0 \\
\textbf{Ours (Mixed Data)} & \textbf{86.0} & \textbf{83.0} \\
\bottomrule
\end{tabular}

\end{table}

\subsection{Exp. 2: Shortcut Learning (Object-Position Coupling)}
\begin{figure}[ht]
    \centering
    \includegraphics[width=0.95\columnwidth]{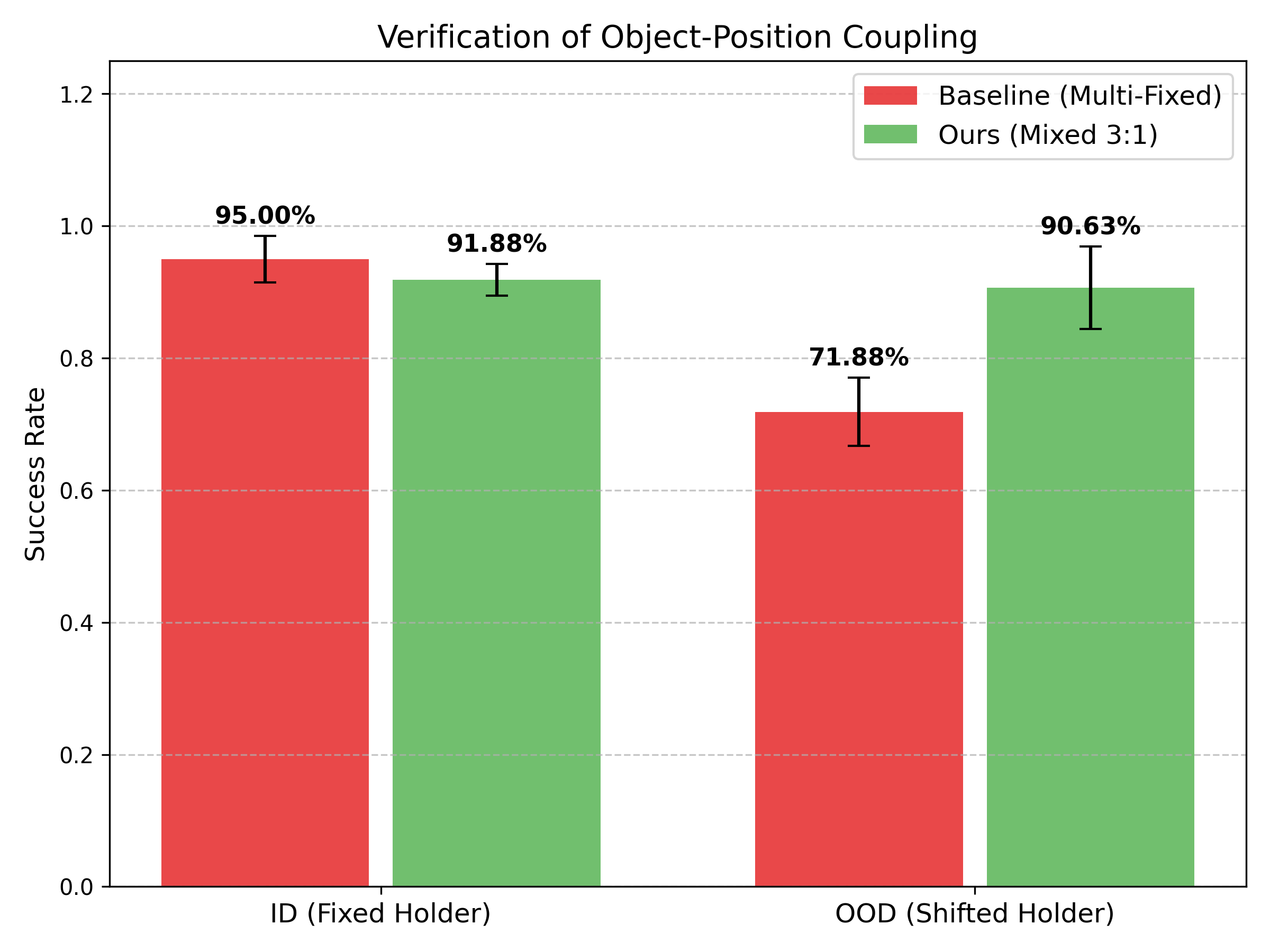}
    \caption{\textbf{Verification of Object-Position Coupling.} The significant drop in performance when the pen holder is shifted (OOD) demonstrates that the model relies on fixed relative positions rather than true spatial understanding. Our method (Green) successfully breaks this coupling.}
    \label{fig:hook}
\end{figure}

To further investigate the mechanism of shortcut learning, we designed a controlled diagnostic experiment (probing setup) to verify \textbf{Object-Position Coupling}: whether the model locks onto the fixed relative position between objects (e.g., pen and holder).
We trained a model on \textbf{Multi-Fixed View} data where, within each group, the camera pose and the pen holder position were fixed and only the pen (target object) was placed at diverse positions. We then tested in two settings:
\begin{itemize}
    \item \textbf{ID-Test (Fixed Holder):} Holder at the same position as in training; pen placed from the same (randomized) set as in training.  %
    \item \textbf{OOD-Test (Shifted Holder):} Holder shifted by one diameter; pen still placed from the same randomized set (only the holder shift is controlled).  %
\end{itemize}
As shown in Fig. \ref{fig:hook} and Table \ref{tab:hook}, the Baseline (Multi-Fixed) achieves a near-perfect \textbf{95.0\%} on ID-Test but drops sharply to \textbf{71.9\%} on OOD-Test.
In contrast, our method (trained with the empirically optimal mix Moving:Multi-Fixed = 1:3) maintains high performance in both settings: \textbf{91.9\%} on ID-Test and \textbf{90.6\%} on OOD-Test.
This confirms that even with diverse camera viewpoints (Multi-Fixed), if the relative position between objects (Object-Position Coupling) is not sufficiently varied, the model does not perfectly capture the true spatial relationship between the pen and the holder. Instead, it learns a spurious correlation (a shortcut) based on the absolute position of the holder or its relation to the camera/base. Our mixed data strategy successfully breaks this coupling, enabling robust generalization.

\begin{table}[h]
\centering
\caption{\textbf{Object-Position Coupling Verification.} Success rates (\%) showing the impact of shifting the object configuration. While the Baseline fails on OOD, our method remains robust.}
\label{tab:hook}
\footnotesize
\begin{tabular}{lcc}
\toprule
\textbf{Test Condition} & \textbf{Baseline} & \textbf{Ours (Mixed 1:3)} \\
\midrule
ID (Fixed Holder) & \textbf{95.0} $\pm$ 3.5 & 91.9 $\pm$ 2.4 \\
OOD (Shifted Holder) & 71.9 $\pm$ 5.2 & \textbf{90.6 $\pm$ 6.3} \\
\bottomrule
\end{tabular}

\end{table}

\subsection{Exp. 3: The Hybrid Dynamic Data and The Golden Ratio of Composition}
To identify a good data composition strategy, we conduct this experiment on the \textbf{Pen Pick-and-Place} task (grasping a pen and inserting it into a pen holder), which allows for rapid policy convergence. We use Gr00t~n1 \cite{bjorck2025gr00t} (hereafter Gr00t) as the primary model; the ratio that maximizes its performance (Moving:Multi-Fixed = 1:3) we call the Golden Ratio below. (Exp.~4 shows that the optimal ratio can vary by architecture.)

\begin{figure}[h]
    \centering
    \includegraphics[width=0.95\columnwidth]{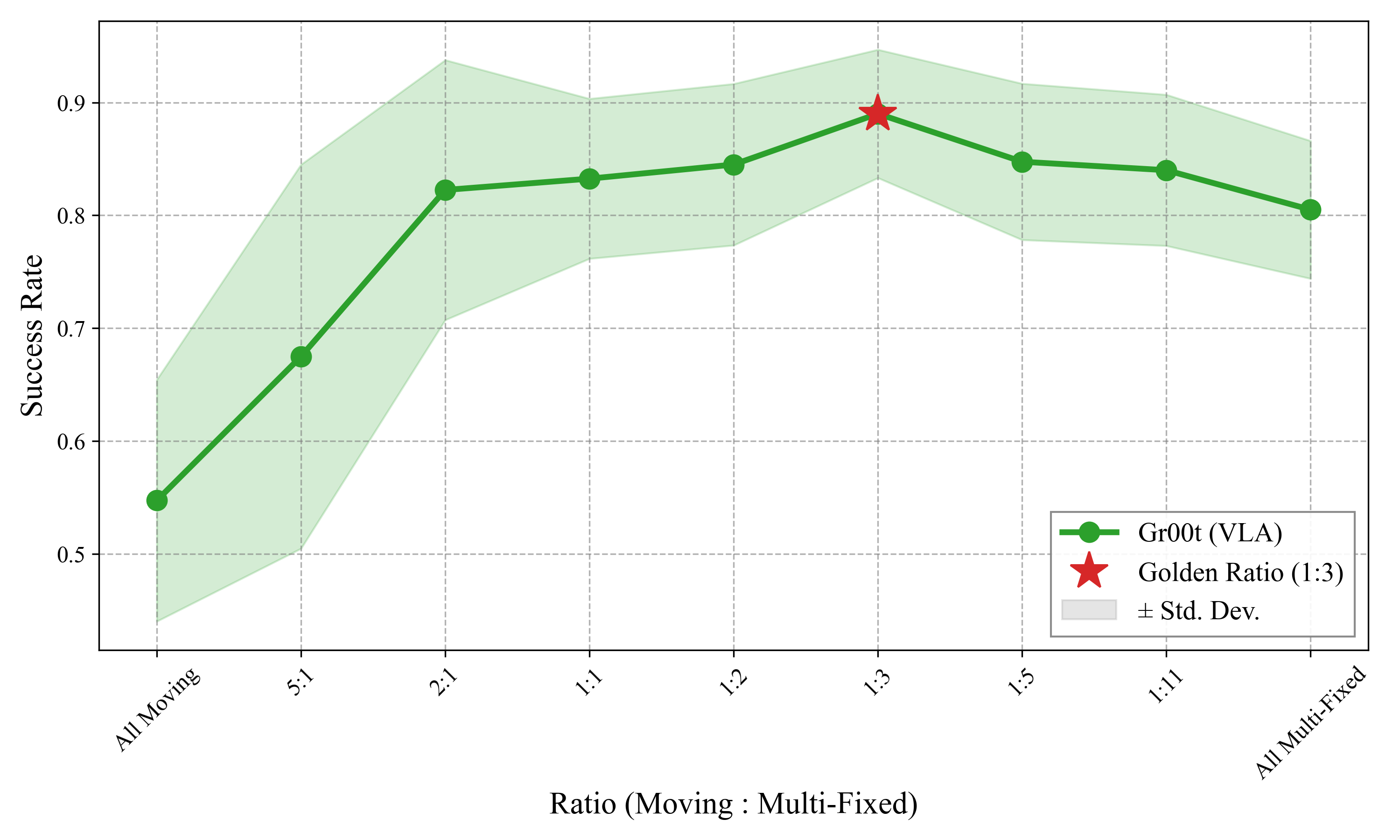}
    \caption{\textbf{Impact of Data Composition Ratio.} For Gr00t, the success rate peaks at Moving:Multi-Fixed = 1:3 (Golden Ratio). Shaded regions denote standard deviation.}
    \label{fig:ratio}
\end{figure}

We investigate the impact of the mixing ratio between Multi-Fixed View and Moving View data. The results reveal a non-trivial insight:
\begin{enumerate}
    \item \textbf{Pure Moving Data is Insufficient (54.8\%):} Although our hypothesis favors Moving View for breaking spurious correlations, training solely on dynamic data performs poorly: the high variance in visual input makes policy convergence difficult.
    \item \textbf{Multi-Fixed Data is a Strong Baseline (80.5\%):} Evaluating all models in this subsection on the \textbf{Moving Test} set, pure Multi-Fixed data reaches 80.5\% (Table~\ref{tab:ratio}), unexpectedly outperforming pure Moving data (54.8\%) on that same test.
    \item \textbf{Mixture is Optimal for Gr00t (89.0\%):} Mixing Multi-Fixed and Moving data further improves the success rate: as shown in Fig. \ref{fig:ratio} and Table \ref{tab:ratio}, for Gr00t a strategic mix of \textbf{Moving:Multi-Fixed = 1:3} yields the best performance.
\end{enumerate}
This confirms that while Multi-Fixed data provides stability for convergence, Moving View data acts as a critical regularizer to enforce spatial invariance.

\begin{table}[h]
\centering
\caption{\textbf{Composition Matters.} Success rates (\%) by Moving:Multi-Fixed ratio. Golden Ratio (1:3) is for Gr00t.}
\label{tab:ratio}
\footnotesize
\begin{tabular}{lcccc}
\toprule
\textbf{Ratio (Moving:Multi-Fixed)} & \textbf{1:0} & \textbf{1:1} & \textbf{1:3} & \textbf{0:1} \\
\midrule
Success Rate (\%) & 54.8 & 83.3 & \textbf{89.0} & 80.5 \\
Std & 10.7 & 7.1 & 5.7 & 6.1 \\
\bottomrule
\end{tabular}

\end{table}

\subsection{Exp. 4: Transfer, Sample Efficiency \& Universality of the Data Strategy}

\subsubsection{Cross-Task Spatial Generalization Transfer}
To evaluate sample efficiency and transferability, we conduct a \textbf{Multi-Object Pick-and-Place} experiment. The multi-task training data consist of \textbf{Fixed View} data only (a single static camera pose; \emph{not} Multi-Fixed) over five object types: watch, lollipop, nail clipper, tape, and cube block.
We compare two settings:
\begin{itemize}
    \item \textbf{Baseline (Fixed Data):} Trained on pure Fixed-View multi-task data (the five objects above).
    \item \textbf{Ours (Mixed Data):} Trained on a mix of 50\% Fixed-View multi-task data and 50\% \textbf{Auxiliary Pen Data} (collected with Moving:Multi-Fixed = 1:3, the Golden Ratio for Gr00t).
\end{itemize}
This setup tests whether the spatial know-how learned from the low-cost auxiliary Pen task (single object type, minimal setup) can robustify a static policy on new tasks.
To reveal the underlying mechanism, we analyze the learned policy via skill decoupling:
\begin{itemize}
    \item \textbf{Semantic Source:} The model learns what to grasp (semantics/affordance) from the Fixed-View Multi-Task data. However, it tends to overfit to the camera pose.
    \item \textbf{Spatial Source:} The model learns ``how to perceive'' (spatial representation/hand-eye coordination) from the Moving-View Pen data, which acts as an auxiliary data source for spatial invariance.
    \item \textbf{Transfer:} During mixed training, the model composes these skills. It uses the spatial perception capability learned from the Pen data to guide the grasping actions learned from the Multi-Task data, enabling it to grasp unseen objects (e.g., watches) under dynamic viewpoints.
\end{itemize}

\begin{figure}[t]
    \centering
    \includegraphics[width=0.95\columnwidth]{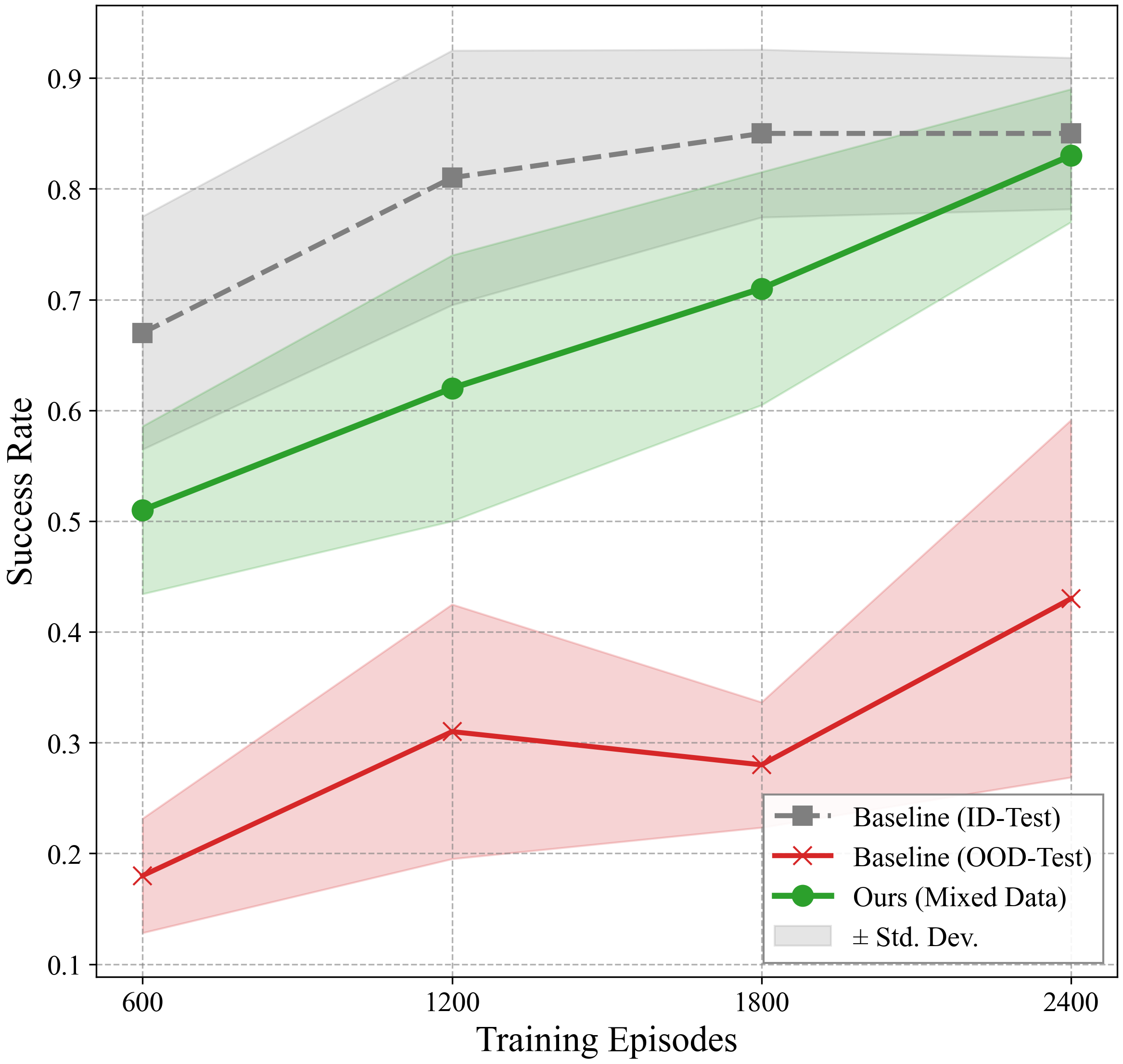}
    \caption{\textbf{Cross-Task Spatial Generalization Transfer.} Our method (Green), leveraging auxiliary data, consistently outperforms the Baseline (Red) on the challenging Moving Test set, achieving high success rates with fewer task-specific episodes. Shaded regions denote standard deviation.}
    \label{fig:efficiency}

\end{figure}

Table \ref{tab:efficiency} and Fig. \ref{fig:efficiency} show that our method consistently achieves higher performance than the baseline on the Moving Test set.

\begin{table}[h]
\centering
\caption{\textbf{Cross-Task Spatial Generalization Transfer.} Success rates (\%) on Moving Test set. Baseline uses pure Fixed-View task data; Ours augments with auxiliary Pen data.}
\label{tab:efficiency}
\resizebox{\columnwidth}{!}{%
\footnotesize
\begin{tabular}{lcccc}
\toprule
\textbf{Episodes} & \textbf{600} & \textbf{1200} & \textbf{1800} & \textbf{2400} \\
\midrule
Baseline (Fixed Data) & 18.0 $\pm$ 5.2 & 31.0 $\pm$ 11.5 & 28.0 $\pm$ 5.7 & 43.0 $\pm$ 16.1 \\
\textbf{Ours (Mixed Data)} & \textbf{51.0 $\pm$ 7.6} & \textbf{62.0 $\pm$ 12.0} & \textbf{71.0 $\pm$ 10.5} & \textbf{83.0 $\pm$ 6.0} \\
\bottomrule
\end{tabular}%
}
\end{table}

\FloatBarrier  %
\subsubsection{Universality of the Data Strategy (Pen Pick-and-Place)}
We revisit the \textbf{Pen Pick-and-Place} task to evaluate different architectures, including ACT \cite{zhao2023learningfinegrainedbimanualmanipulation}, Diffusion \cite{chi2023diffusionpolicy}, and Pi0 ($\pi_{0.5}$) \cite{intelligence2025pi05visionlanguageactionmodelopenworld}. We compare their performance using mix (Moving:Multi-Fixed = 1:1, 1:3, 1:11) versus pure (Multi-Fixed, Moving) data. Note that we conduct basic tuning but not exhaustive optimization on the ACT, Diffusion, and Pi0 models, as our focus is to demonstrate that the data strategy benefits all architectures rather than to compare model capacity.

\begin{figure}[t]
    \centering
    \includegraphics[width=0.95\columnwidth]{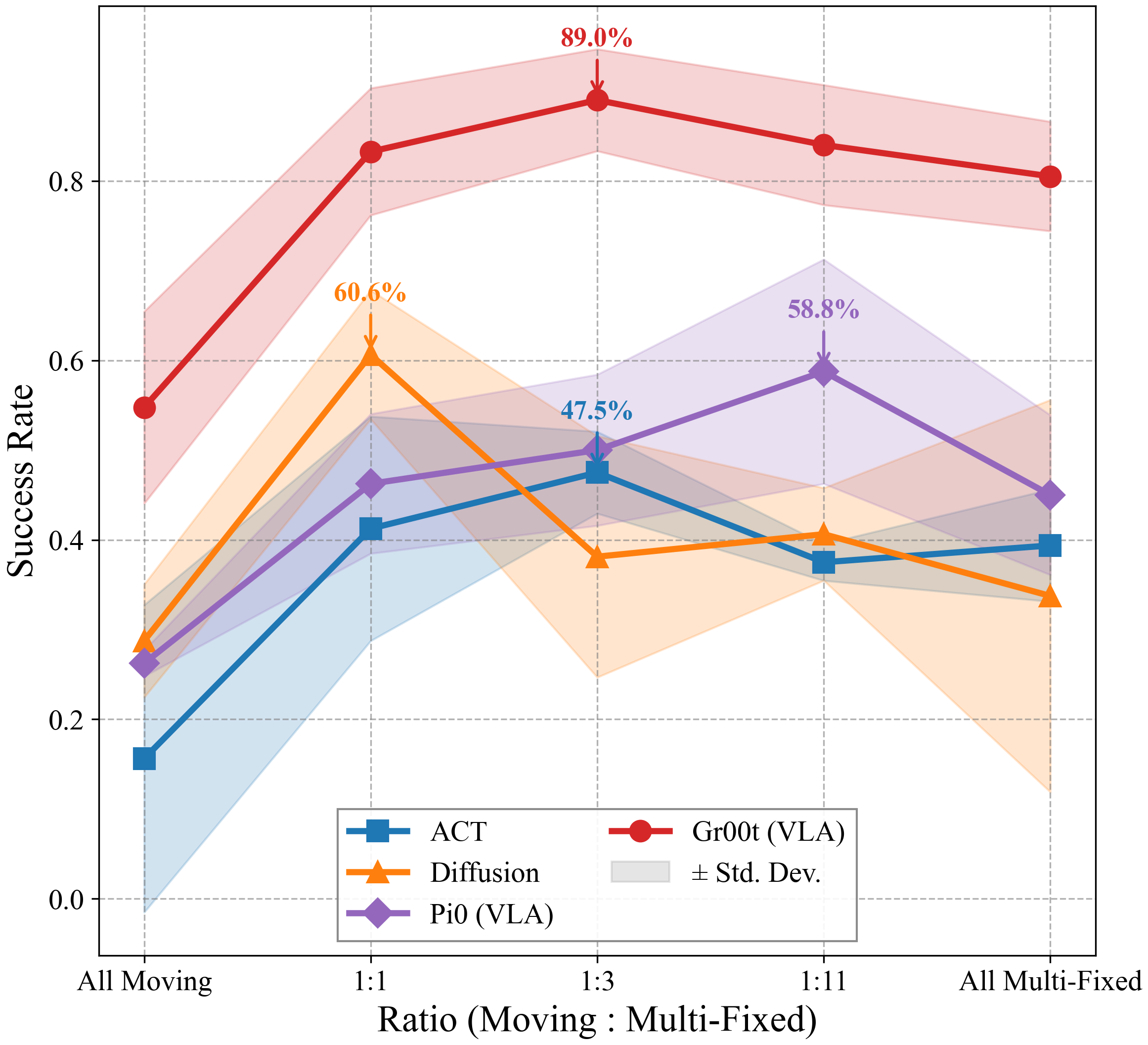}
    \caption{\textbf{Universality of the Data Strategy.} All evaluated architectures (ACT, Diffusion, and VLA models including Pi0 and Gr00t) benefit from the mixed data strategy, with Gr00t showing the highest peak performance. Shaded regions denote standard deviation.}
    \label{fig:models}
\end{figure}

As shown in Fig. \ref{fig:models} and Table \ref{tab:models}, the convergence difficulty with pure Moving data is consistently observed across architectures, while Multi-Fixed data yields uniformly better performance. Our results suggest that it is essential to strike a balance between learning efficiency and data diversity, where the latter introduces little to no degradation given sufficient data.

Furthermore, the benefits of our proposed Hybrid Dynamic Data Collection strategy are universal across architectures. Although the optimal Moving:Multi-Fixed ratio varies (e.g., 1:1 for Diffusion, 1:3 for Gr00t), all models achieve substantial performance gains (e.g., Diffusion +26.8\%, Pi0 +13.8\%) by incorporating Moving-View data.

\begin{table}[h]
\centering
\caption{\textbf{Universal Benefit.} All models improve with Mixed Data. Best Mix denotes performance at the optimal ratio for each model.}
\label{tab:models}
\footnotesize
\begin{tabular}{lcccc}
\toprule
\textbf{Model} & \textbf{Multi-Fixed} & \textbf{Best Mix} & \textbf{Ratio} & \textbf{Gain} \\
\midrule
ACT & 39.4 $\pm$ 6.3 & 47.5 $\pm$ 4.6 & 1:3 & +8.1 \\
Diffusion & 33.8 $\pm$ 21.8 & 60.6 $\pm$ 7.2 & 1:1 & \textbf{+26.8} \\
Pi0 & 45.0 $\pm$ 8.9 & 58.8 $\pm$ 12.5 & 1:11 & +13.8 \\
\textbf{Gr00t (VLA)} & \textbf{80.5 $\pm$ 6.1} & \textbf{89.0 $\pm$ 5.7} & \textbf{1:3} & +8.5 \\
\bottomrule
\end{tabular}

\end{table}

\section{CONCLUSIONS}

In summary, we provide systematic real-world validation for the proposed Hybrid Dynamic Data Collection system. We demonstrate that spatial generalization requires more than just moving the camera; it requires breaking implicit couplings. By implementing our hybrid dynamic data strategy in the real world, we enable VLAs to mitigate the pitfall of shortcut learning. Our experiments demonstrate that Multi-Fixed data ensures stable convergence, whereas Moving View data enhances data diversity, with their combination achieving the best results, and this beneficial pattern holds universally across different architectures. Furthermore, we demonstrate that the core manipulation capability under moving views learned from hybrid dynamic pen pick-and-place data can be robustly transferred to new tasks, with improved sample efficiency as a favorable side effect.

\textbf{Limitations and Future Work.} Our evaluation focuses on table-top pick-and-place (a five-category multi-object task) and contact-rich alignment (pen insertion); the data strategy is task-agnostic and we expect transfer to longer-horizon, contact-rich tasks, though systematic validation remains future work. Methodologically, a more fundamental step is to lift moving viewpoints into an online module \emph{inside} the VLA---policy-conditioned active viewpoint control (a learned next-best-view) that resolves occlusion while retaining the viewpoint invariance our data provides.

\bibliographystyle{IEEEtran}
\setlength{\bibitemsep}{0pt} 
\balance 
\bibliography{references}

\end{document}